\documentclass{article}
\usepackage{spconf,amsmath,amssymb,graphicx}
\usepackage{todonotes}
\usepackage[hidelinks]{hyperref}
\usepackage{multirow}
\usepackage{xcolor, makecell, array, booktabs}
\usepackage{siunitx}
\usepackage[inline]{enumitem}

\title{Investigating Modality Bias in Audio Visual Video Parsing}
\name{Piyush Singh Pasi, Shubham Nemani, Preethi Jyothi, Ganesh Ramakrishnan}

\address{Indian Institute of Technology, Bombay}
\begin{document}
\ninept
\maketitle
\begin{abstract}
We focus on the audio-visual video parsing (AVVP) problem that involves detecting audio and visual event labels with temporal boundaries. The task is especially challenging since it is weakly supervised with only event labels available as a bag of labels for each video. An existing  state-of-the-art model for AVVP uses a hybrid attention network (HAN) to generate cross-modal features for both audio and visual modalities, and an attentive pooling module that aggregates predicted audio and visual segment-level event probabilities to yield video-level event probabilities. We provide a detailed analysis of modality bias in the existing HAN architecture, where a modality is completely ignored during prediction. We also propose a variant of feature aggregation in HAN that leads to an absolute gain in F-scores of about $2\%$ and $1.6\%$ for visual and audio-visual events at both segment-level and event-level, in comparison to the existing HAN model. 
\end{abstract}
\begin{keywords}
AVVP, weakly-supervised learning, modality bias
\end{keywords}
\section{Introduction}
The Audio-Visual Video Parsing (AVVP)~\cite{tian2020avvp} task involves the fine-grained parsing of a video to generate temporal audio, video and audio-visual event labels. The additional challenge in this task is that there is only weak supervision during training in the form of event labels for the entire video, whereas the objective is to predict fine-grained audio and visual events for temporal event segments. AVVP is regarded as a Multimodal Multiple Instance Learning (MMIL) problem and it has a number of applications in audio-visual source separation and other video understanding tasks.

In order to utilize the weak labels effectively, Tian {\em et. al.}~\cite{tian2020avvp} proposed a Hybrid Attention Network (HAN) that generates aggregated feature representations for the audio and visual modalities with attention within and across modalities. These aggregated representations are further processed by an attentive MMIL pooling module that combines the representations using attention weights to produce a probability distribution across event labels for each video. This probability distribution can be directly used within a cross-entropy loss with the weak video-level labels serving as ground-truth.

After carefully analyzing the training losses and the assumptions in the model proposed  by~\cite{tian2020avvp}, we revisit their design choice of using cross attention for both modalities when constructing aggregated features. We hypothesize that using cross-modal attention on the audio modality but only self-attention on the visual modality could be more effective. And we indeed  empirically verify that this variant outperforms the model in \cite{tian2020avvp} with significant gains on the visual evaluation metrics.

When learning from multiple modalities, the model could establish spurious correlations between the target event and one (or more) modalities and not extract meaningful signals from all modalities. Such spurious correlations could lead to reasonable downstream task performance on the training dataset. However, the resulting models would demonstrate a modality bias and would not generalize well to test instances where the ignored modalities have rich signals. Tian {\em et. al.}~\cite{tian2020avvp} aim at alleviating modality bias by added modality-specific losses and label smoothing. Our experiments reveal that label smoothing could in fact reinforce modality biases depending on which modalities use smoothed labels, and hence should be used carefully.

Our main contributions can be summarized as follows:
\begin{enumerate}
    \item We propose a simple and well-motivated variant of feature aggregation within the Hybrid Attention Network that yields significant gains on the evaluation metrics compared to~\cite{tian2020avvp}.
    \item We carefully analyze the effect of label smoothing and explain how it causes drastic shifts in the audio-visual attention distributions generated by attentive MMIL pooling. We also examine its impact on the final performance using a detailed ablation study \footnote{More details at: \url{https://www.cse.iitb.ac.in/~malta/mbias}}.
\end{enumerate}.

\section{Related Work}

\subsection{Audio-Visual Representation Learning}
Videos contain information in both audio and visual modalities. Due to temporal synchronization in audio and visual data, most works~\cite{listen2look, look_listen_attend, robust_asr, rouditchenko2020avlnet, korbar2018cooperative, aytar2016soundnet, owens2016ambient} focus on learning joint embedding to exploit information present in both the modalities. Ying \emph{ et. al.}~\cite{look_listen_attend} employ cross-attention and self-attention for audio-visual representation learning. Ruohan \emph{et. al.}~\cite{listen2look} propose an attention based LSTM network for fusing audio-visual information and use it for action recognition. George \emph{et. al.}~\cite{robust_asr} use cross-modal alignment for robust speech recognition. The Hybrid Attention Network HAN~\cite{tian2020avvp} jointly models the audio and visual modalities via self-attention and cross-attention.

\subsection{Multiple Instance Learning (MIL)}
Multiple instance learning (MIL)~\cite{mil, chen2006miles, zhang2001dd, babenko2009visual, ilse2018attention, maron1998multiple} is a form of supervised learning in which the dataset consists of bags. Each bag contains multiple training instances and we have labels for a bag; each instance in the bag is itself unlabeled. From a collection of labeled bags, the learner tries to induce a model that will label individual instances correctly. 
Yapeng \emph{et. al.} in \cite{mil_event} formulate audio-visual event localization as a MIL problem. They use an audio-visual snippet pair as a single instance, in contrast to AVVP in which the audio and visual snippet at the same time instance are considered as two separate instances. Similarly, multiple instance learning has also been used for object detection~\cite{cmil, domain_mil}. 

\section{AVVP}\label{avvp}
\begin{figure}[t]
  \centering
  \includegraphics[width=0.9\linewidth]{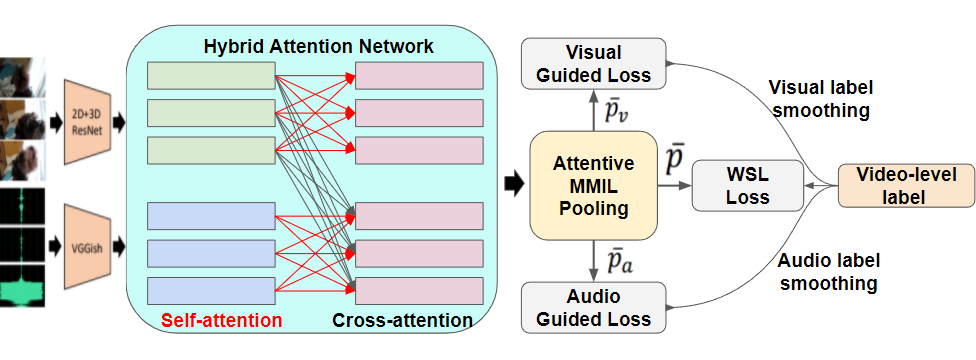}
  \caption{Depiction of architectural changes made to the HAN network~\cite{tian2020avvp}. The block in the aqua colour highlights modifications made to the
attention module for $A_{cross}$, $V_{self}$ (there is no cross attention for visual features).}
  \label{fig:HAN ACVS}
\end{figure}
As described in~\cite{tian2020avvp}, each video in the AVVP task is divided into $T$ segments of audio and visual snippet pairs denoted by $\{A_t, V_t\}_{t=1}^T$. Each snippet $\{A_t, V_t\}$ is associated with an event label set $\boldsymbol{y}_t = \{y^a_t, y^v_t, y^{av}_t\}$; here, $y^a_t, y^v_t, y^{av}_t$ are vectors of dimensionality $C$ denoting audio, visual and audio-visual event labels respectively, where $C$ is the size of the event label set. Only video-level labels are assumed to be available during training (resulting from weak supervision), while audio and visual events will be predicted for each video snippet during inference.

\cite{tian2020avvp} proposed the use of a hybrid attention network (HAN) and attentive multimodal multiple instance learning (MMIL) pooling for the AVVP task. First, the audio and visual snippet pairs $\{A_t, V_t\}_{t=1}^T$ are passed through pretrained feature extractors to generate audio and visual representations $\{f_a^t\}_{t=1}^T$ and $\{f_v^t\}_{t=1}^T$, respectively. Next, HAN learns hybrid attention functions from both audio and visual representations at each time-step $t$, $\boldsymbol{f}_a = [f_a^1, f_a^2,..., f_a^T]$ and $\boldsymbol{f}_v = [f_v^1, f_v^2,..., f_v^T]$, to yield the following aggregate audio ($A_{cross}$ or  $\hat{f_a^t}$) and video ($V_{cross}$ or  $\hat{f_v^t}$)  representations:
\begin{align}
    A_{cross} \equiv \hat{f_a^t} = f_a^t + g_{sa}(f_a^t, \boldsymbol{f}_a) + g_{ca}(f_a^t, \boldsymbol{f}_v)   \label{eq:han1} \\
    V_{cross} \equiv \hat{f_v^t} = f_v^t + g_{sa}(f_v^t, \boldsymbol{f}_v) + g_{ca}(f_v^t, \boldsymbol{f}_a)
    \label{eq:han2}
\end{align}
where $g_{sa}$ and $g_{ca}$ are self-attention and cross-modal attention functions, respectively. These attention functions capture intra-modal and inter-modal similarities and are computed using the standard attention formulation~\cite{transformer}: 
$g_{sa}(f_a^t, \boldsymbol{f}_a) = \mathrm{softmax}({f_a^t \boldsymbol{f}_a^{\intercal}}/{\sqrt{d}})\boldsymbol{f}_a$ and $g_{ca}(f_a^t, \boldsymbol{f}_v) = \mathrm{softmax}({f_a^t \boldsymbol{f}_v^{\intercal}}/{\sqrt{d}})\boldsymbol{f}_v$, 
where $d$ is the dimensionality of the audio/visual features.
\vspace{0.5em}

\noindent \textbf{Attentive MMIL Pooling.}
\cite{tian2020avvp} uses attentive MMIL pooling to leverage weakly-supervised video-level labels during training. In \cite{tian2020avvp}, the aggregated temporal features $\{\hat{f_a^t}, \hat{f_v^t}\}_{t=1}^T$ are passed through a shared fully-connected (FC) layer with sigmoid activation to obtain output probabilities for each individual event category.
The predicted audio and visual event probabilities for time-step $t$ are $p_a^t$ and $p_v^t$, respectively.
The audio and visual event probabilities are aggregated using attentive MMIL pooling to predict the video-level event probability $\bar{\boldsymbol{p}}_{wsl}$ as follows:
\begin{equation}
    \bar{\boldsymbol{p}}_{wsl} = \sum_{t=1}^{T} \sum_{m=1}^{M} (W_{tp} \odot W_{av} \odot P) [t, m, :]
    \label{eq:9}
\end{equation}
where $P(t,1,:) = p_a^t$ and $P(t,2,:) = p_v^t$, $\odot$ denotes element-wise multiplication, $M$ is the total number of modalities ({\em i.e.}, 2 in AVVP where $m \in \{1,2\}$ refers to the audio and visual modalities, respectively). $W_{tp}$ and $W_{av}$ are temporal attention and audio-visual attention tensors predicted from the aggregated features $\{\hat{f_a^t}, \hat{f_v^t}\}_{t=1}^T$ respectively. $W_{tp}$ and $W_{av}$ are computed as $W_{tp}[:, m, c] = \mathrm{softmax}(F_{tp}[:,m,c])$ and $W_{av}[t, :, c] = \mathrm{softmax}(F_{av}[t,:,c])$ where $F_{tp}$ and $F_{av}$ refer to two different fully-connected layers, $t=1,2,\ldots,T$, $m=1,2$ and $c=1,2,\ldots,C$. The ground truth labels $\bar{\boldsymbol{y}}$ and the predicted video-level event probabilities $\bar{\boldsymbol{p}}_{wsl}$ are used to optimize the binary cross-entropy loss function specified as: $\mathcal{L}_{wsl} = CE(\bar{\boldsymbol{p}}_{wsl}, \bar{\boldsymbol{y}}) = -\sum_{c=1}^{C} \bar{\boldsymbol{y}}[c]log(\bar{\boldsymbol{p}}_{wsl}[c])$.
For better results and to alleviate modality bias, \cite{tian2020avvp} proposes the use of cross-entropy losses specific to each individual modality. This modality-guided loss, $\mathcal{L}_g$ is computed as follows: $\mathcal{L}_{g} = \mathcal{L}_{a} + \mathcal{L}_{v} = CE(\boldsymbol{\bar{p}}_a, \boldsymbol{\bar{y}}_a) + CE(\boldsymbol{\bar{p}}_v, \boldsymbol{\bar{y}}_v)$, where  
$\boldsymbol{\bar{y}}_a$ and $\boldsymbol{\bar{y}}_v$ are video-level ground truth labels for the audio and visual modalities, respectively (that is simply initialized as $\boldsymbol{\bar{y}}_a = \boldsymbol{\bar{y}}_v = \boldsymbol{\bar{y}}$),  $\boldsymbol{\bar{p}}_a$ and $\boldsymbol{\bar{p}}_v$ are video-level audio and visual event probabilities given by:
\begin{equation}
    \boldsymbol{\bar{p}}_a = \sum_{t=1}^T (W_{tp} \odot P)[t,1,:]
    \label{eq:12}
\end{equation}
\begin{equation}
    \boldsymbol{\bar{p}}_v = \sum_{t=1}^T (W_{tp} \odot P)[t,2,:]
    \label{eq:13}
\end{equation}
During training, the model is optimized over the combined loss $\mathcal{L} = \mathcal{L}_{wsl}+\mathcal{L}_{g}$.
\vspace{0.5em}

\noindent \textbf{Label Smoothing.}
Since we do not have video-level ground truth labels for each modality, $\boldsymbol{\bar{y}}_a$ and $\boldsymbol{\bar{y}}_v$ are set to  $\boldsymbol{\bar{y}}$ which lacks modality-specific information. Tian {\em et. al.}~\cite{tian2020avvp} suggest using label smoothing on $\boldsymbol{\bar{y}}_a$ and $\boldsymbol{\bar{y}}_v$, as some events in $\boldsymbol{\bar{y}}$ can be noise for a particular modality. For example, if $\boldsymbol{\bar{y}}$ = \{\textit{Telephone\_bell\_ringing, Cat}\}, the telephone bell could be ringing in the background without any accompanying visual cues and conversely, the cat could only be seen in the video and not heard at all.  Label smoothing is implemented in \cite{tian2020avvp} as: $\boldsymbol{\bar{y}}_m = (1-\delta_m)\boldsymbol{\bar{y}} + \frac{\delta_m}{K}$ where $m \in \{a,v\}$ for audio and visual modalities; $\delta_m \in [0,1)$ is the confidence parameter yielding a convex combination of (i) the event probability distribution and (ii) the uniform distribution $U = \frac{1}{K}$ (where $K > 1$) which helps in smoothing the event probability distribution.

\begin{figure}[t]
  \centering
  \includegraphics[width=0.9\linewidth]{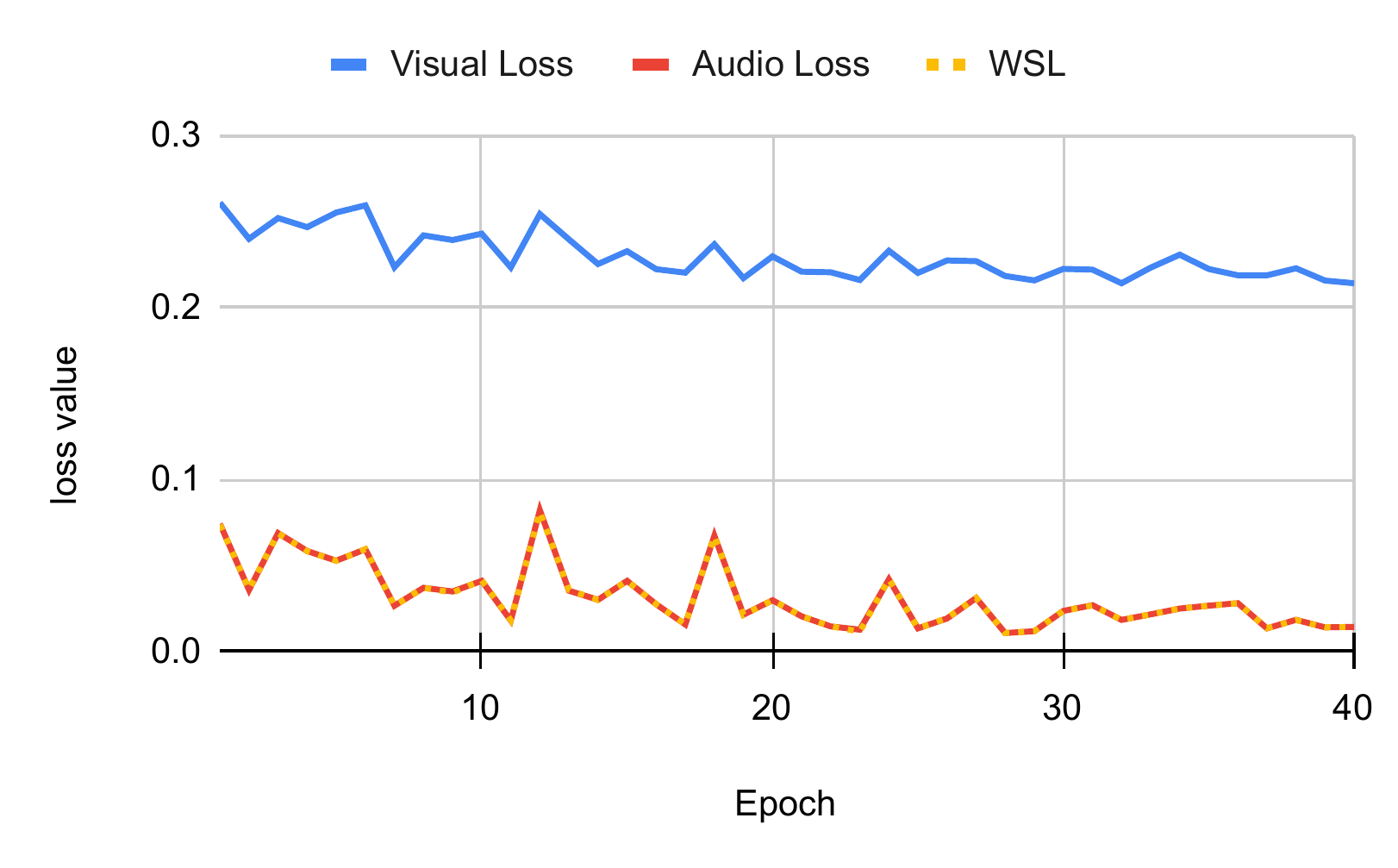}
  \caption{Comparison between loss values of visual loss $\mathcal{L}_{v}$, audio loss $\mathcal{L}_{a}$, and weakly-supervised loss $\mathcal{L}_{wsl}$ for HAN. \cite{tian2020avvp}}
  \label{fig:loss_comparison}
\end{figure}

\vspace{0.5 em}
\noindent \textbf{Behaviour of Training Losses.} 
Figure~\ref{fig:loss_comparison} shows the loss values corresponding to $\mathcal{L}_{a}$, $\mathcal{L}_{v}$, and $\mathcal{L}_{wsl}$ during training.
We observe that the weakly-supervised loss $\mathcal{L}_{wsl}$ and the audio loss functions $\mathcal{L}_{a}$ nearly coincide. The mean squared difference between $\mathcal{L}_{wsl}$ and $\mathcal{L}_{a}$ for 40 epochs denoted by $MSE_{wsl,a}$ is 9e-8. In contrast, the mean squared difference between $\mathcal{L}_{wsl}$ and $\mathcal{L}_{v}$ denoted by $MSE_{wsl,v}$ is 0.104. 
This curious training artefact of $\mathcal{L}_{wsl}$ and  $\mathcal{L}_{a}$ mirroring each other can be explained as follows. 

Given that $\mathcal{L}_{wsl}$ is a binary cross-entropy loss function between $\boldsymbol{\bar{p}}_{wsl}$ and $\boldsymbol{\bar{y}}$ and $\mathcal{L}_{a}$ is a binary cross-entropy loss function of $\boldsymbol{\bar{p}}_a$ and $\boldsymbol{\bar{y}}$, the only difference appears in the event probability values. For $\mathcal{L}_{wsl}$ to imitate $\mathcal{L}_{a}$, the video-level event probability values $\boldsymbol{\bar{p}}_{wsl}$ and the audio event probabilities $\boldsymbol{\bar{p}}_a$ should be very close to each other. 
From equations~\ref{eq:9}, \ref{eq:12}, and \ref{eq:13}, $\boldsymbol{\bar{p}}_{wsl}$ can be written as a weighted combination of $\boldsymbol{\bar{p}}_a$ and $\boldsymbol{\bar{p}_v}$:
\begin{equation}
    \bar{\boldsymbol{p}}_{wsl} = W_a \odot \bar{\boldsymbol{p}}_a + W_v \odot \bar{\boldsymbol{p}}_v
    \label{eq:16}
\end{equation}
where $W_a = W_{av}[:, 1, :]$ and $W_v = W_{av}[:, 2, :]$ correspond to audio and visual attention tensors, respectively. If $\boldsymbol{p}_{wsl}$ and $\boldsymbol{p}_{a}$ become close to each, then $W_a \approx \boldsymbol{1}$ and $W_v \approx \boldsymbol{0}$. This could be attributed to label smoothing (which will be discussed in detail in Section~\ref{analysis}). This modality bias also motivates us to explore different variants of the aggregate features that we outline in the next section.

\section{Proposed Variants of Aggregate Features}
\label{Variants_of_Aggregate_Features}
Tian {\em et. al.}~\cite{tian2020avvp} propose using the HAN network to generate aggregate audio and video representations, {\em viz.}, $\hat{f_a^t}$ and $\hat{f_v^t}$, each containing both self-attention and cross-attention based functions. However, this symmetric treatment of constructing audio and video representations may not be the best choice due to the nature of the modalities involved. Consider the example of $\boldsymbol{\bar{y}}_m = \{\textit{Telephone\_bell\_ringing, Fire\_alarm}\}$. These events can occur in the background without any supporting visual clues, and hence will not appear in the ground truth event labels for the visual modality. However, cross-attention from audio in this case could induce false signals of the presence of these events in the visual modality and subsequently hurt performance. Audio events, on the other hand, could benefit from the visual modality in helping disambiguate between audio sounds ({\em e.g.}, $\{\textit{Blender, Vacuum\_cleaner}\}$) using visual clues. These observations suggest that {\em a model with visual features aggregated using only self-attention and audio features aggregated using both self-attention and cross-attention might perform better.} 

We remove the cross-modal attention function $g_{ca}()$ while computing aggregated features for the visual modality to get:
\begin{equation}
    V_{self} = \tilde{f_v^t} = g(f_v^t, \boldsymbol{f}_a,  \boldsymbol{f}_v) = f_v^t + g_{sa}(f_v^t, \boldsymbol{f}_v)
    \label{eq:5}
\end{equation}
%
For a complete analysis, we can similarly remove cross-attention from the aggregated features for the audio modality to get: 
\begin{equation}
    A_{self} = \tilde{f_a^t} = g(f_a^t, \boldsymbol{f}_v,  \boldsymbol{f}_a) = f_a^t + g_{sa}(f_a^t, \boldsymbol{f}_a)
    \label{eq:6}
\end{equation}
%
Recall that  in Eq.~\eqref{eq:han1} and~\eqref{eq:han2}, we referred to the aggregated audio and visual features as $A_{cross}$ and $V_{cross}$, respectively.

The model $A_{cross} + V_{self}$ will refer to the above-mentioned feature aggregates where the visual features do not use cross-attention from audio. For the sake of completeness, we also compare against the remaining three variants, $A_{self} + V_{self}$, $A_{self} + V_{cross}$ and $A_{cross} + V_{cross}$. Note that $A_{cross} + V_{cross}$ is the same model proposed in~\cite{tian2020avvp}.

\begin{figure}[t]
  \centering
  \includegraphics[width=0.95\linewidth]{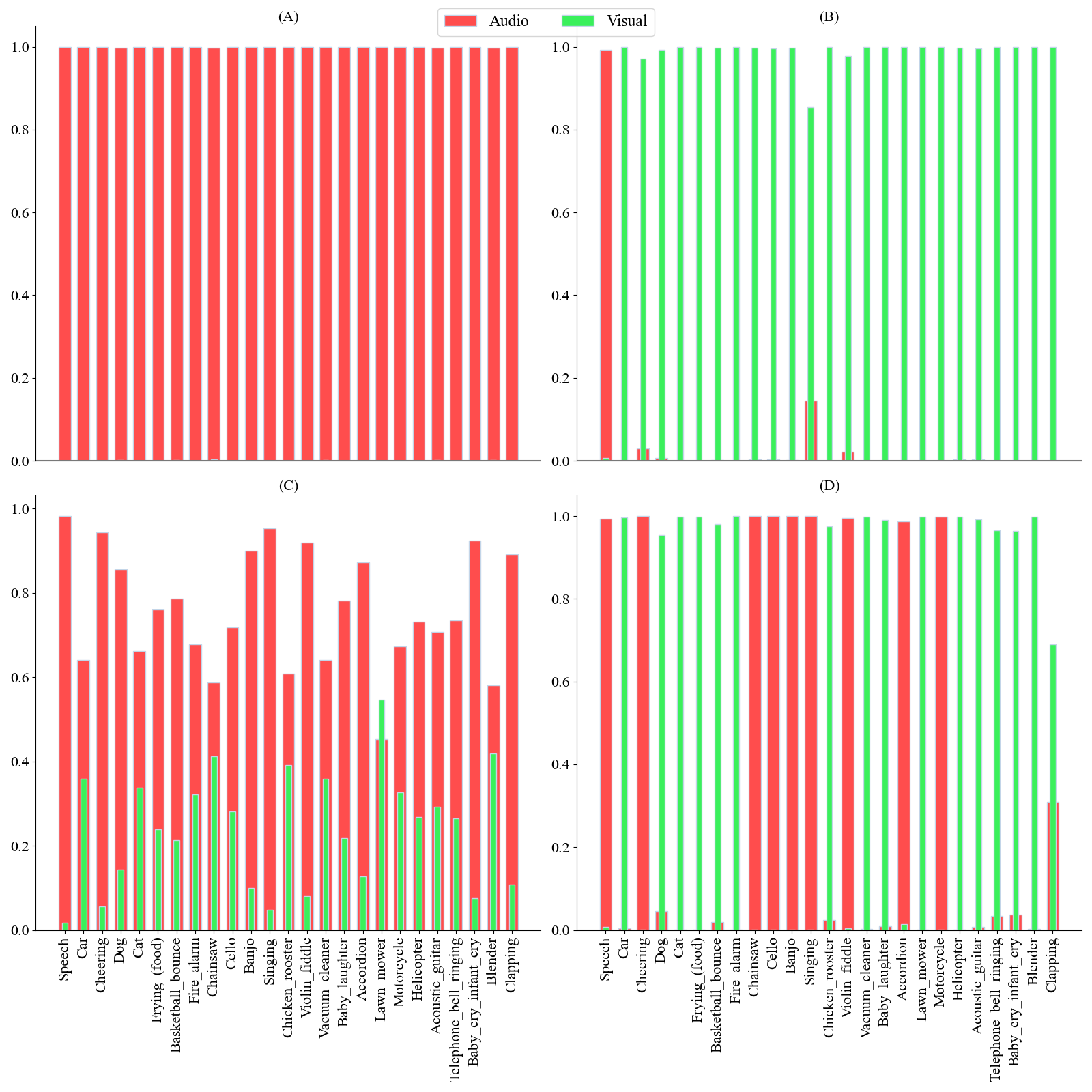}
  \caption{Category-wise audio-visual attention weights distribution aggregated over the test set for different types of label smoothing. Visual(A), Audio(B), no smoothing(C) and Audio-Visual(D)}
  \label{fig:av attn}
  
\end{figure}

\section{Experiments}

\subsection{Experimental Setup}The LLP dataset~\cite{tian2020avvp} contains 11,849 YouTube video clips with 25 event categories for a total of 32.9 hours. Each video is  10-second-long. The training set consists of 10,000 videos with weak labels i.e video-level event labels. The validation and test sets (that are manually annotated with segment-level labels) have 649 and 1200 videos, respectively. Following~\cite{tian2020avvp}, we sample these videos at 8 frames per second and break into non-overlapping segments of length 1 second. Features extracted from ResNet152~\cite{resNet152} and 3D ResNet~\cite{resNet3D} are fused to get 512-dimensional segment-level visual features. We use a VGGish extractor~\cite{VGGish} to extract 128-dimensional segment-level audio features. We train our model for 40 epochs with batch size 16, Adam as the optimizer, and  initial learning rate set to 3e-4. The learning rate is dropped by a factor of 0.1 after every 10 epochs.

\subsection{Evaluation Metrics}
All our models are evaluated on
the metrics proposed by~\cite{tian2020avvp}. We use F-scores for audio, visual, and audio-visual events for segment-level and event-level metrics, indicating segment-level performance and  video-level performance, respectively. To calculate event-level metrics, consecutive events of the same category are concatenated, and the F-score is calculated based on \textit{mIoU} = 0.5 as the threshold. The Ty@AV metric refers to an average over audio, visual, and audio-visual evaluation results. The Ev@AV metric calculates the F-score for all audio and visual events of each video.

\begin{table}[t]
  \label{tab:1}
  \centering
  \begin{tabular}{l|c|c c}
    \toprule
    Event type & Method & \thead{Segment\\level} &\thead{Event\\level}\\
    \midrule
    \multirow{4}{*}{Audio}   &  $A_{self}+V_{self}$   & 52.1  & 41.6  \\
            & $A_{self}+V_{cross}$ & 49.9  & 39.2  \\ 
            & $A_{cross}+V_{self}$ & \textbf{60.5}  & \textbf{51.9}  \\
            & $A_{cross}+V_{cross}$ \cite{tian2020avvp}  & 60.1    & 51.3 \\\midrule
    \multirow{4}{*}{Visual}  & $A_{self}+V_{self}$ & 53.4	& 48.7 \\
            & $A_{self}+V_{cross}$ & 53.9	& 49  \\ 
            & $A_{cross}+V_{self}$ & \textbf{54.9}	& \textbf{51.2} \\
            & $A_{cross}+V_{cross}$ \cite{tian2020avvp}   & 52.9	& 48.9 \\\midrule
    \multirow{4}{*}{\thead{Audio\\\&\\Visual}}    & $A_{self}+V_{self}$ & 41.2	& 33.4  \\
            & $A_{self}+V_{cross}$ & 40	& 30.9  \\ 
            & $A_{cross}+V_{self}$ & \textbf{50.5}	& \textbf{44.3}  \\
            & $A_{cross}+V_{cross}$ \cite{tian2020avvp}  & 48.9	& 43 \\\midrule
    \multirow{4}{*}{Ty@AV} &$A_{self}+V_{self}$ & 48.9	& 41.2 \\
            & $A_{self}+V_{cross}$ & 48	& 39.7  \\  
            &$A_{cross}+V_{self}$& \textbf{55.3}	& \textbf{49.1}  \\
            & $A_{cross}+V_{cross}$ \cite{tian2020avvp}   & 54	& 47.7 \\\midrule
    \multirow{4}{*}{Ev@AV}    & $A_{self}+V_{self}$  & 52.2	& 42.7  \\
            & $A_{self}+V_{cross}$  & 51.5	& 42.3  \\ 
            & $A_{cross}+V_{self}$ & \textbf{56.5}	& \textbf{48.9}  \\
            &$A_{cross}+V_{cross}$ \cite{tian2020avvp}   & 55.4	& 48 \\
    \bottomrule
  \end{tabular}
  \caption{Results of AVVP on LLP dataset for variants of HAN}
\end{table}

\begin{table}[t]
  \label{tab:2}
  \centering
  \resizebox{\columnwidth}{!}{%
  \begin{tabular}{c|c|c c|c c}
    \toprule
     \multirow{3}{*}{\thead{Event\\type}} & \multirow{3}{*}{\thead{Smoothing\\modality}} &  
       \multicolumn{2}{c|}{$A_{cross} + V_{cross}$\cite{tian2020avvp}} &
     \multicolumn{2}{c}{$A_{cross} + V_{self}$}  \\
   & & \thead{Segment\\level} &\thead{Event\\level} & \thead{Segment\\level} &\thead{Event\\level}  \\
    \midrule
   \multirow{4}{*}{Audio}   & No-LS   & 58  & 49.7  & 60.3  & 52.1 \\
            & LS-A  & 57.9  & 49.1   & 60.3  & 51.8 \\ 
            & LS-V  & \textbf{60.1}  & \textbf{51.3}  & \textbf{60.5}  & \textbf{51.9}  \\
            & LS-AV    & 57.5    & 48  & 59.9    & 51  \\\midrule
    \multirow{4}{*}{Visual}  & No-LS  & 52.6	& 48.6  & 53.7	& 50 \\
            & LS-A & 53.1	& 48.5  & 53.7	& 50.4  \\ 
            & LS-V  & 52.9	& 48.9  & \textbf{54.9}	& \textbf{51.2}  \\
            & LS-AV  & \textbf{54.3}	& \textbf{50.3}  & 49.3	& 42.8  \\\midrule
    \multirow{4}{*}{\thead{Audio\\ \& \\Visual}}    & No-LS & 47.6	& 41.4  & 49.4	& 43.8   \\
            & LS-A & 47.7	& 41 & 49.4	& 43.8 \\ 
            & LS-V  & \textbf{48.9}	& \textbf{43}   & \textbf{50.5}	& \textbf{44.3}   \\
            & LS-AV   & 48.6	& 42.2  & 49.3	& 42.8 \\\midrule
    \multirow{4}{*}{Ty@AV} & No-LS  & 52.7	& 46.6  & 54.5	& 48.7 \\
            & LS-A & 52.9	& 46.2 & 54.5	& 48.7  \\  
            & LS-V & \textbf{54}	& \textbf{47.7} & \textbf{55.3}	& \textbf{49.1}  \\
            & LS-AV   & 53.4	& 46.8  & 54.3	& 47.7 \\\midrule
    \multirow{4}{*}{Ev@AV}    & No-LS  & 54.3	& 47.3  & 56.3	& 48.7  \\
            & LS-A  & 54.9	& 47  & 56	& 48.8  \\ 
            & LS-V & \textbf{55.4}	& \textbf{48}  & \textbf{56.5}	& \textbf{48.9}  \\
            &  LS-AV    & 54.7	& 46.6  & 56	& 47.8  \\
    \bottomrule
  \end{tabular}}
  \caption{Effect of Label Smoothing on HAN \cite{tian2020avvp}  i.e. $A_{cross} + V_{cross}$ and $A_{cross} + V_{self}$. No-LS implies no label smoothing; LS-A denotes smoothing only in audio modality; LS-V denotes smoothing only in visual modality; LS-AV denotes smoothing on both audio and visual modalities.}
\end{table}

\subsection{Results and Analysis}\label{analysis}

Table~\ref{tab:1} shows the segment and event-level F1-scores for AVVP using the four different variants of aggregate features detailed in Section~\ref{Variants_of_Aggregate_Features}. We make the following two key observations. \begin{enumerate*}
    \item $A_{cross}+ V_{self}$ is the best performing variant across all five evaluation metrics, and consistently outperforms the baseline model~\cite{tian2020avvp}, {\em viz.}, $A_{cross}+ V_{cross}$. 
    \item Using $A_{self}$ instead of $A_{cross}$ leads to a large and consistent drop in performance across all metrics. This suggests that the audio modality clearly benefits from cross-modal attention from the visual modality.
\end{enumerate*} 

\vspace{0.5em}
\noindent \textbf{Label Smoothing.} 
Figure \ref{fig:av attn} shows the audio-visual attention weight distributions aggregated for the test set using four different types of label smoothing, {\em viz.}, (A) Smoothing of labels of only the audio (LS-A) or (B) only of video (LS-V), (C) no smoothing at all (No-LS) and (D) smoothing the labels of both modalities (LS-AV). It is evident that smoothing only the labels of one modality ({\em i.e.}, visual in plot A and audio in plot B) leads to the attention weights being completely biased towards the other modality ({\em i.e.}, audio in plot A and visual in plot B). Removing label smoothing entirely or adding label smoothing to both modalities yields attention distributions without a clear modality bias (shown in (C) and (D)). 

This behaviour can be explained by examining the effect of label smoothing on the losses and its subsequent effect on the audio-visual attention weights. Adding smoothing to a particular modality $m \in \{a,v\}$ makes $\boldsymbol{\bar{y}}_m$ a real-valued vector, increasing the terms in $\mathcal{L}_{m}$ to the total number of events. In the case of multi-hot vectors, only a few terms equal to the number of ground-truth events are present. This poses a challenge to minimizing the loss. In LS-A, $\boldsymbol{\bar{y}}, \boldsymbol{\bar{y}}_v$ are multi-hot vectors and $\boldsymbol{\bar{y}}_a$ is a real-valued vector. Setting $W_a = \boldsymbol{0}$ and $W_v = \boldsymbol{1}$ will yield $\boldsymbol{\bar{p}}_{wsl}$ in the form of a multi-hot vector and will thus minimize $\mathcal{L}_{wsl}$. Any other $W_a$ and $W_v$ will not generate a multi-hot vector. Similarly, in LS-V, the model picks $W_a = \boldsymbol{1}$ and $W_v = \boldsymbol{0}$ to minimize $\mathcal{L}_{wsl}$.
In LS-AV, for $\boldsymbol{\bar{p}}_{wsl}$ to be closest to the multi-hot vector form, in the absence of an event, setting $W_m = \boldsymbol{1}$ for modality with least probability is the best resort for $\mathcal{L}_{wsl}$. Similarly, in the presence of an event, setting $W_m = \boldsymbol{1}$ for modality with the highest probability will minimize $\mathcal{L}_{wsl}$ more. When there is no label smoothing, such extreme skewness vanishes.

Table \ref{tab:2} shows the segment-level and event-level results using $A_{cross}+V_{self}$ and $A_{cross}+V_{cross}$ with four types of label smoothing (LS-A, LS-V, No-LS and LS-AV). When compared to the model No-LS, the model LS-V shows a significant increase in the performance of the audio modality. Similarly, label smoothing on audio modality {\em i.e.}, LS-A shows a drop in audio evaluation metrics and some gain on visual evaluation metrics. On applying label smoothing on both modalities {\em i.e.}, LS-AV, the model favors the visual modality more. This aligns with the audio-visual attention weights in Fig.~\ref{fig:av attn} (D). Another interesting observation from Table~\ref{tab:2} is that $A_{cross}+V_{self}$ is more robust to label smoothing compared to $A_{cross}+V_{cross}$; the averaged variance of F1-scores of audio, visual and audio-visual events across types of label smoothing is $0.35$ for $A_{cross}+V_{self}$ compared to $0.95$ for $A_{cross}+V_{cross}$. 
Such empirical stability, particularly in the segment-level evaluation metrics, is owing to our reformulated HAN- $A_{cross}+V_{self}$ and aligns with the intuition stated in Section \ref{Variants_of_Aggregate_Features}. 
Audio events can occur in the background with no support for visual cues in the frames. Using such events as ground truth in $\mathcal{L}_v$ reduces the performance, since the ground truth itself is noisy, and thus label smoothing helps. 

\section{Conclusions}
In this work, we focus on the AVVP problem and specifically study the issue of modality bias in the main model proposed for this task in~\cite{tian2020avvp}. We trace the source of modality bias to label smoothing that was a part of the originally proposed framework for AVVP. We propose a new variant for aggregating features within this framework that is not only more accurate than the baseline but is also more robust to label smoothing. As part of future work, we propose to develop modality-aware techniques that explicitly discourage  modality bias in the model objective.

\bibliographystyle{IEEEbib}
\bibliography{refs}

\end{document}